# nnDetection for Intracranial Aneurysms Detection and Localization

Maysam Orouskhani[1], Negar Firoozeh[1], Shaojun Xia[2], Mahmud Mossa-Basha[1], Chengcheng Zhu[1]

[1]Department of Radiology, University of Washington
[2]Peking University Cancer Hospital Institute, Beijing, China
maysam@uw.edu

**Abstract**. Intracranial aneurysms are a commonly occurring and life-threatening condition, affecting approximately 3.2% of the general population. Consequently, the detection of these aneurysms plays a crucial role in their management. Lesion detection involves the simultaneous localization and categorization of abnormalities within medical images. In this particular study, we employed the nnDetection framework, a self-configuring framework specifically designed for 3D medical object detection, to effectively detect and localize the 3D coordinates of aneurysms. To capture and extract diverse features associated with aneurysms, we utilized two modalities: TOF-MRA and structural MRI, both obtained from the ADAM dataset. The performance of our proposed deep learning model was assessed through the utilization of free-response receiver operative characteristics for evaluation purposes. The model's weights and 3D prediction of the bounding box of TOF-MRA are publicly available at https://github.com/orouskhani/AneurysmDetection.

## 1  Introduction

An intracranial aneurysm, also known as a cerebral aneurysm or brain aneurysm, refers to a weak or thin area in the wall of a blood vessel within the brain. This weakened section of the blood vessel forms a bulge or ballooning, similar to a bubble. If left untreated, an intracranial aneurysm can rupture, leading to a potentially life-threatening condition called a subarachnoid hemorrhage. Intracranial aneurysms are responsible for approximately 80%–90% of nontraumatic subarachnoid hemorrhages, leading to mortality rates ranging 23%–51% and a 10%–20% risk of permanent disability [1]. Therefore, early detection of intracranial aneurysms has significant implications for the diagnosis, treatment, and prognosis of patients, as well as the role of radiologists in managing these conditions. When an aneurysm is detected at an early stage, it allows for timely intervention and reduces the risk of rupture, which can lead to severe consequences such as hemorrhagic stroke. Furthermore, early detection also facilitates the monitoring and follow-up of patients, allowing for timely adjustments in the treatment plan if necessary. Radiologists play a crucial role in the detection and characterization of intracranial aneurysms, utilizing their expertise in interpreting imaging studies to identify these vascular abnormalities. Their accurate and timely diagnosis aids in the prompt initiation of treatment, leading to improved patient outcomes and overall prognosis. However, with the advent of deep learning techniques, there has been a growing interest in leveraging artificial intelligence algorithms to improve the accuracy and efficiency of aneurysm detection.

These advancements have the potential to revolutionize aneurysm diagnosis and management, providing clinicians with valuable guidance in iden-tifying and monitoring aneurysms, ultimately enhancing patient outcomes, and improving overall healthcare delivery in the field of neurovascular medicine. Re-cently nnU-Net [2] as a self-configuring deep learning framework was proposed with an automatic hyperparameter tuning tool for medical image segmentation. Building on this, nnDetection [3] has been introduced to automate the process and configuration of medical object detection.

In our study, we leverage the nnDetection deep learning model to detect and localize intracranial aneurysms. The proposed model demonstrates signif-icant advantages compared to previous methods, as it simultaneously localizes and categorizes aneurysms within each image slice without requiring manual in-tervention. Moreover, it automatically generates bounding boxes and provides estimated 3D coordinates for each aneurysm, enabling precise localization. These features enhance the model's usability and clinical relevance in aneurysm man-agement.

The main contribution of this work is as follows:

– We used both modalities of ADAM dataset including TOF-MRA and struc-tural MRI to localize the 3D coordination of intracranial aneurysms in the brain.
– After aneurysms localization, we generated the bounding box including the aneurysms.

## 2  Method

In this study, we utilized the ADAM dataset, which comprises 113 TOF-MRA and structural MRI scans obtained from 93 patients diagnosed with unruptured intracranial aneurysms (UIAs) [4]. The dataset encompassed a total of 125 UIAs, and two radiologists manually annotated the voxel-wise locations on the axial plane. The MRIs were conducted at the UMC Utrecht, the Netherlands, using Philips scanners with either 1.5 or 3T field strength. The TOF-MRAs had vary-ing in-plane resolutions ranging from 0.2 to 1 mm, and the slice thickness fell within the range of 0.4 to 0.7 mm. Notably, there was no standardized acquisi-tion protocol for the TOF-MRAs. Figure 1 showcases two representative cases extracted from the ADAM dataset.

To automate the detection and localization of intracranial aneurysms within the ADAM dataset, we employed a 3D full resolution nnDetection deep learning model. This model utilizes the Retina U-net architecture [5], which combines the Retina Net detector with the widely recognized U-Net segmentation model. The Retina Net component, acting as a one-shot detector, directly performs the classification and bounding box regression tasks by leveraging intermediate activation maps from the Feature Pyramid Network (FPN) decoder blocks [5].

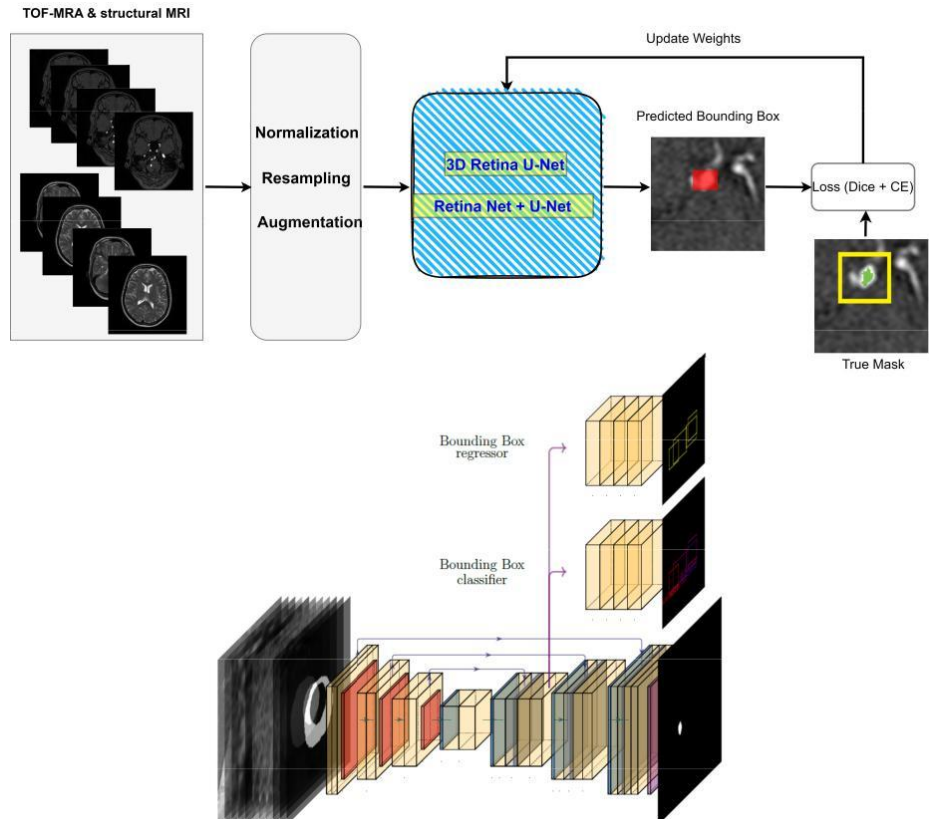

Fig. 1. Up: The proposed nnDetection deep learning model for the aneurysm detection task. The nnDetection fully automatically detects and localizes the aneurysm by generating a bounding box using 3D Retina U-Net. Down: Overview of the Retina U-Net architecture. On the bottom, a U-Net-like architecture segments the lesions present in the image irrespective of their class. On the top, a bounding box regression head takes a feature map from a decoder of the U-Net and refines the coarse detections, while the bounding box classifier tries to predict their class. These two heads visit all decoder levels, performing detection at different scales transparently. [8]

Figure 2 provides an overview of the detection model and the Retina U-net architecture. In terms of loss functions, the Retina U-Net employs pixel-wise cross-entropy loss and soft Dice loss [5] for segmentation, while binary cross-entropy (BCE) and generalized intersection over union (GIoU) [7] are employed for classification and box regression, respectively. Similar to nnU-Net, nnDetection incorporates pre-processing techniques such as cropping, Z-Score normalization, and scaling. The patch size is adjusted while adapting the network architecture, and the batch size is fixed at four. The encoder of the Retina U-Net comprises plain convolutions, ReLU activation, and instance normalization blocks. The model was trained for 100 epochs with 2500 mini-batches per epoch, and five-fold cross-validation was applied. Stochastic Gradient Descent (SGD) with Nesterov momentum of 0.9 was utilized to update the weights. All models were trained using three RTX 3090 GPUs, and the patch size was set to 256 × 224 × 56.

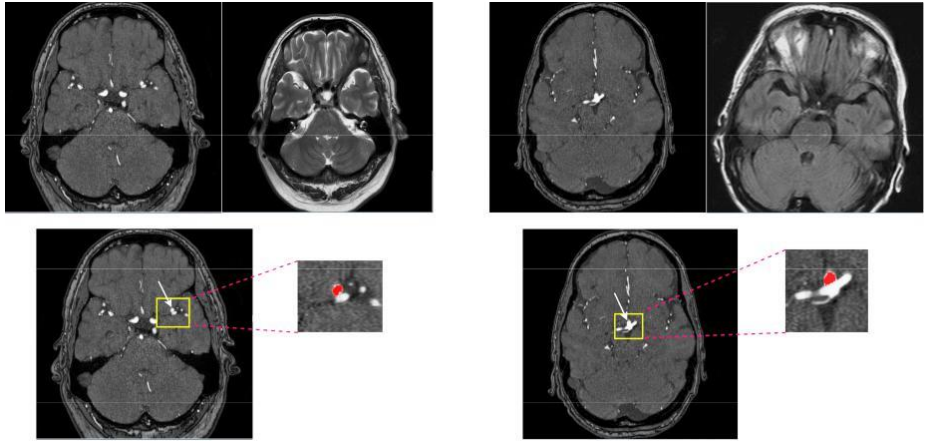

Fig. 2. . Left: case 22, aneurysm 3D coordination: (353, 230, 78). Right: case 78, aneurysm 3D coordination: (268, 262, 54) from ADAM dataset

## 3    Experiments

### 3.1   Data

ADAM composed of 113 TOF-MRA and structural MRI (93 patients with unruptured intracranial aneurysm [UIAs]). The total number of UIAs was 125 and the voxel-wise annotations were drawn in the axial plane by two radiologists.

### 3.2   Results

To evaluate the performance of the deep learning model in the aneurysm detection task, the free-response operating characteristic (FROC) measure was plotted. The FROC calculates the lesion-level sensitivity versus false positive per scan (FPPS). The FROC curve is like ROC analysis, except that the false positive rate on the x-axis is replaced by the average number of false positives per image.

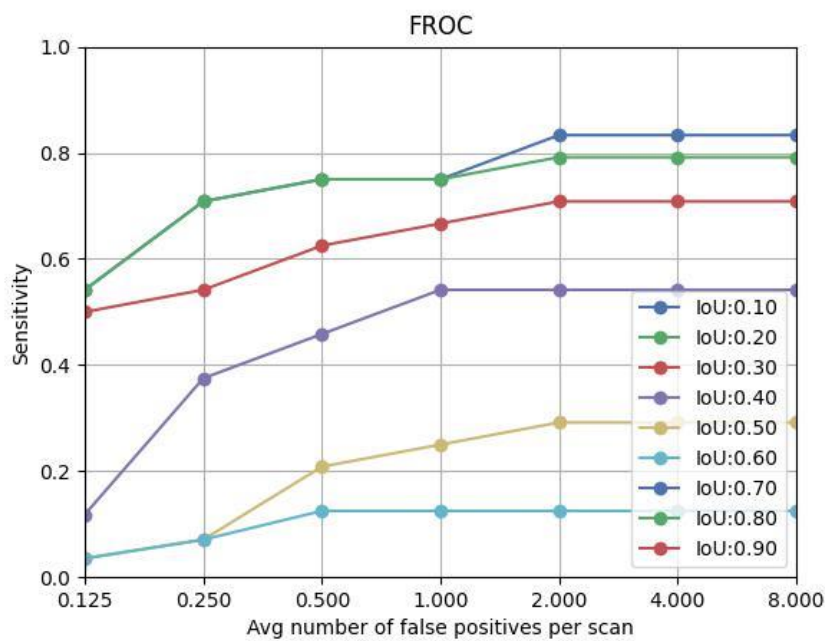

Fig. 3. The FROC curve: the measure of sensitivity as vertical axis vs the average number of false positive per scan as horizontal axis

As an example, using the threshold of IoU as 0.3, at FPPS = 0.25, 0.5, 1, 2, the sensitivity was 55%, 62%, 68%, and 72%. In addition to the bounding box, which was generated to show the location of the aneurysm, the 3D coordination of aneurysms (center) was calculated to indicate the exact location of the aneurysm in each case. Figure 4 (A) showed the comparison of the bounding box predicted by our model with the true segmentation mask in four test cases. The estimated values of the center of the aneurysm as a 3D coordination are also compared with the exact values provided by the ADAM dataset in Figure 4(B).

## 4 Conclusion

This study presented a deep learning-based approach for the detection and localization of intracranial aneurysms using TOF-MRA and structural MRI data. The utilization of the nnDetection framework, incorporating the Retina U-Net architecture, enabled accurate identification and localization of aneurysms within the medical images. The model effectively drew bounding boxes around the aneurysms in each image slice, providing precise localization information. Evaluation of the deep learning model was performed using the FROC measure, demonstrating its robust performance in detecting aneurysms. Additionally, the estimated coordinates of the aneurysms were compared against the ground truth masks, further validating the accuracy of the localization predictions. This study contributes to the field of automated aneurysm detection and localization, offering potential advancements in clinical decision-making and management strategies for intracranial aneurysms.

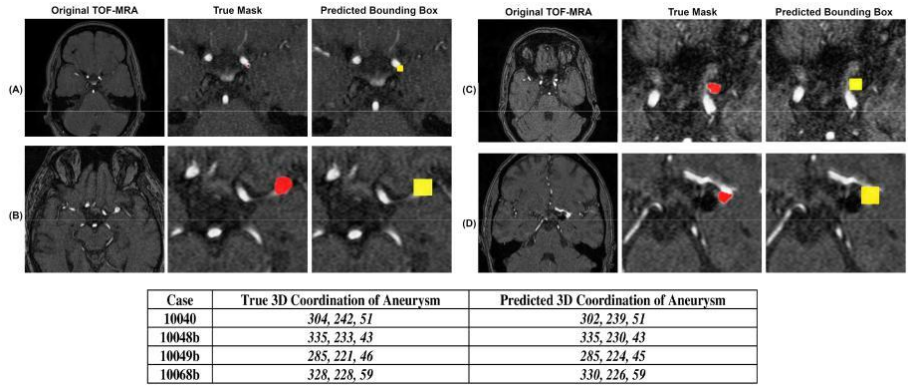

Fig. 4. The results of aneurysm detection by nnDetection. The generated bounding box including aneurysm for cases as follows: A) case 10040, B) case 10048, C) case 10049, D) case 10068. The table also shows the true 3D center of mass voxel coordinates x,y,z and compares with the estimated coordination